\documentclass[pdftex]{article}

\usepackage[final, nonatbib]{neurips_2023}
\usepackage[square,sort,comma,numbers]{natbib}
\usepackage{lipsum}
\usepackage{algorithm}
\usepackage{multirow}
\usepackage{algpseudocode}
\usepackage{amsmath}
\usepackage{amssymb}
\usepackage{natbib}
\usepackage{soul}
\usepackage{algorithm}
\usepackage{adjustbox}

\usepackage[utf8]{inputenc} 
\usepackage[T1]{fontenc}    
\usepackage{hyperref}       
\usepackage{url}            
\usepackage{booktabs}       
\usepackage{amsfonts}       
\usepackage{nicefrac}       
\usepackage{microtype}      
\usepackage{xcolor}         
\usepackage[pdftex]{graphicx} 
\usepackage{microtype}
\usepackage{subfigure}
\usepackage{booktabs} 
\usepackage{color,colortbl}
\usepackage[linewidth=1pt]{mdframed}
\usepackage{framed}
\usepackage{adjustbox}

\usepackage{amsmath}
\usepackage{amssymb}
\usepackage{mathtools}
\usepackage{amsthm}

\usepackage[capitalize,noabbrev]{cleveref}

\theoremstyle{plain}

\theoremstyle{definition}

\theoremstyle{remark}

\usepackage[textsize=tiny]{todonotes}

\title{ZNorm: Z-Score Gradient Normalization Accelerating Skip-Connected Network Training without Architectural Modification}

\author{
  Juyoung Yun$^{1,2}$\thanks{Corresponding Author: \texttt{juyoung.yun@stonybrook.edu}. This work has been accepted to the 39th Annual AAAI Conference on Artificial Intelligence 2025 Workshop: The first Workshop on Scalable and Efficient Artificial Intelligence Systems (SEAS)} \\ \\
  $^1$Stony Brook University, Department of Computer Science, USA\\
  $^2$ Stony Brook University, Department of Applied Mathematics and Statistics, USA\\
}

\begin{document}

\maketitle

\let\thefootnote\relax\footnotetext{}

\begin{abstract}
The rapid advancements in deep learning necessitate better training methods for deep neural networks (DNNs). As models grow in complexity, vanishing and exploding gradients impede performance, particularly in skip-connected architectures like Deep Residual Networks. We propose Z-Score Normalization for Gradient Descent (ZNorm), an innovative technique that adjusts only the gradients without modifying the network architecture to accelerate training and improve model performance. ZNorm normalizes the overall gradients, providing consistent gradient scaling across layers, effectively reducing the risks of vanishing and exploding gradients and achieving superior performance. Extensive experiments on CIFAR-10 and medical datasets confirm that ZNorm consistently outperforms existing methods under the same experimental settings. In medical imaging applications, ZNorm significantly enhances tumor prediction and segmentation accuracy, underscoring its practical utility. These findings highlight ZNorm’s potential as a robust and versatile tool for enhancing the training and effectiveness of deep neural networks, especially in skip-connected architectures, across various applications.
\end{abstract}

\section{Introduction}

Efficient training of deep neural networks (DNNs) has become increasingly critical as deep learning models continue to grow in depth and complexity\cite{LeCun2015}. However, training DNNs remains a challenging task due to issues such as vanishing and exploding gradients, which limit performance and scalability\cite{Pascanu2013}. Recently, Vision Transformers (ViTs)~\cite{dosovitskiy2020image} have gained popularity for their impressive performance in computer vision tasks. However, convolutional neural networks (CNNs), especially skip-connected networks like ResNet, DenseNet, and U-Net, remain widely used due to their effectiveness and robustness, particularly in scenarios with smaller datasets~\cite{small} or specific tasks like medical imaging~\cite{util1,util2,util3}. Ensuring efficient training of skip-connected networks is crucial, as these architectures continue to play a significant role in a variety of applications.

Numerous methods have been developed to address gradient-related challenges~\cite{effic}, focusing primarily on normalizing activations or weights to stabilize training. Techniques such as Batch Normalization (BN)\cite{ioffe2015batch}, Layer Normalization (LN)\cite{ba2016layer}, and Weight Standardization (WS)\cite{qiao2019weight} have become standard in many architectures. While these methods improve training stability, they often introduce architectural modifications, increasing implementation complexity and reducing flexibility in certain scenarios. Gradient-based techniques such as Gradient Centralization\cite{center}, Gradient Clipping\cite{pascanu2013difficulty}, and Weight Decay\cite{krogh1991simple} demonstrate that focusing on gradients alone can yield significant performance improvements without altering network structures. 

Building upon this principle, we propose Z-score Gradient Normalization (ZNorm), a novel approach to efficient training that directly adjusts gradients to ensure consistent scaling across layers. By normalizing both the mean and variance of gradients, where stable gradient flow is critical to achieving optimal performance. Our core contributions in this paper are as follows: 

\begin{itemize} 
\item We improve model performance by adjusting gradients only, without modifying neural network architecture. 
\item ZNorm is simple to implement, making it adaptable to various models with skip-connection. 
\end{itemize} 

ZNorm exemplifies efficient training by accelerating training and improving model performance without requiring architectural changes. Unlike activation-based normalization techniques, ZNorm operates exclusively on gradients, making it a lightweight yet powerful tool for optimizing skip-connected networks like ResNet and U-Net. Its compatibility with these architectures enhances their robustness against vanishing and exploding gradients, enabling consistent and efficient training.

To demonstrate its effectiveness, we evaluate ZNorm on a variety of skip-connected networks across synthetic and real-world datasets. Experiments show that ZNorm consistently outperforms existing methods, achieving superior performance in tasks such as image classification and medical imaging applications. In particular, ZNorm significantly enhances tumor prediction and segmentation accuracy, showcasing its utility in high-impact scenarios. These findings establish ZNorm as a robust and versatile tool for efficient training of skip-connected networks, capable of improving both training speed and model performance across diverse applications.

\section{Related Works}

Optimization techniques are central to training deep neural networks effectively. Stochastic gradient descent (SGD) and its variants, such as SGD with momentum \citep{qian1999momentum} and adaptive methods like Adam \citep{kingma2014adam}, have been crucial in optimizing complex architectures. These methods aim to minimize the loss function via parameter updates. Adaptive optimizers like Adagrad \citep{duchi2011adaptive} and RMSProp \citep{tieleman2012lecture} further adjust learning rates dynamically based on gradient magnitude and history, stabilizing the training process for deep models. AdamW \citep{adamw} improves upon traditional weight decay by decoupling L2 regularization from the gradient update in adaptive optimization methods like Adam. In traditional Adam, weight decay is applied implicitly through the L2 regularization term, which affects both the gradient and the parameter updates. AdamW, however, applies weight decay directly to the weights during the parameter update step, ensuring that the regularization only affects the weight magnitudes and not the gradient direction. The gradient update for weight decay in AdamW is modified as $\nabla_\theta L_{\text{decay}}(\theta_t) = \nabla_\theta L(\theta_t) + \lambda \theta_t$ where $\lambda$ is the weight decay coefficient. This decoupled approach in AdamW prevents the regularization term from interfering with the adaptive learning rate updates, leading to more stable training and better generalization, especially in large-scale models. 

Training deep neural networks (DNNs) effectively requires overcoming challenges such as vanishing and exploding gradients, which hinder the performance of complex models~\citep{Pascanu2013}. Traditional approaches to address these issues include normalization techniques applied to activations and weights, such as Batch Normalization (BN)~\citep{ioffe2015batch}, Layer Normalization (LN)~\citep{ba2016layer}, and Weight Standardization (WS)~\citep{qiao2019weight}. While effective, these techniques require modifications to the network architecture, which can add complexity and limit flexibility. Additionally, large-batch optimization techniques, such as LARS~\citep{lamb} and LAMB~\citep{lars}, provide layer-wise adaptive learning rates, enabling stable training with large batches. While these methods effectively scale learning rates based on weight norms, they are primarily designed for large-batch settings and may not generalize to smaller batch sizes or different architectures. In contrast, ZNorm enhances training stability through direct gradient normalization, eliminating the need for complex adjustments or specific batch configurations. Unlike batch-dependent methods, ZNorm performs normalization on the gradient tensors themselves, ensuring consistent behavior regardless of the chosen batch size. 

Gradient adjustment methods directly manipulate the gradient updates to improve training stability and prevent issues like exploding or vanishing gradients. Gradient Clipping~\citep{pascanu2013difficulty} is one such method that caps gradient magnitudes to a threshold $\tau$, effectively preventing the exploding gradient problem. If the gradient exceeds $\tau$, it is rescaled like $\nabla_\theta L(\theta_t) \leftarrow \nabla_\theta L(\theta_t) \cdot \min\left(1, \frac{\tau}{||\nabla_\theta L(\theta_t)||}\right)$. While Gradient Clipping is effective for stabilizing training, it lacks comprehensive control over gradient distribution across layers, potentially leading to inconsistencies in gradient scaling. Gradient Centralization (GC)~\citep{center} is another method that normalizes gradients by zero-centering them, effectively enhancing training consistency and performance. GC achieves this by adjusting the gradient mean to zero $\nabla_\theta L(\theta_t) \leftarrow \nabla_\theta L(\theta_t) - \frac{1}{n} \sum_{i=1}^{n} \nabla_{\theta_i} L(\theta_t)$ where $n$ is the number of gradient elements. Although GC improves generalization and speeds up convergence, it only adjusts the mean, leaving the variance uncontrolled, which may not fully stabilize training in deep networks. SING~\citep{courtois2023sing} is another recent approach, offering plug-and-play gradient standardization to improve stability with optimizers such as AdamW. SING normalizes gradients layer-wise, enhancing generalization and allowing escape from narrow local minima. However, SING’s layer-wise adjustments are dependent on specific optimizers and introduce additional hyperparameters, whereas ZNorm operates independently of optimizer choice and requires no extra tuning, making it simpler to implement.

Building on these approaches, our work introduces Z-score Gradient Normalization (ZNorm), a technique that standardizes both the mean and variance of gradients across layers, ensuring consistent scaling. By normalizing the gradients using Z-scores, ZNorm directly addresses both vanishing and exploding gradients, particularly enhancing stability in skip-connected architectures like ResNet and U-Net. Unlike GC, which only zero-centers the gradient, ZNorm performs full normalization $\nabla_\theta L(\theta_t) \leftarrow \frac{\nabla_\theta L(\theta_t) - \mu_{\nabla}}{\sigma_{\nabla}}$ where $\mu_{\nabla}$ and $\sigma_{\nabla}$ are the mean and standard deviation of the gradient elements, respectively. This additional variance control provides more stable training across diverse skip-connected architectures and applications.

In summary, ZNorm positions itself as a straightforward and adaptable gradient normalization technique that avoids architectural modifications. By focusing solely on gradient adjustments, ZNorm can be easily integrated into various training processes with minimal changes. This layer-wise gradient scaling achieves stable and efficient training, particularly enhancing performance in skip-connected networks such as ResNet, DenseNet, and U-Net. As a result, ZNorm proves to be a valuable tool for improving training speed and effectiveness across diverse applications.

Furthermore, by only adjusting the gradients, ZNorm simplifies the training pipeline, making it suitable for rapid implementation without the need for extensive reconfiguration or additional hyperparameters. This adaptability allows ZNorm to be applied flexibly across different network architectures and tasks, ensuring that stable training benefits are accessible without requiring fundamental changes to the underlying model design.

\section{Methodology}
In this section we introduces Z-Score Normalization (ZNorm), a technique designed to standardize gradient tensors across all network layers. \\

\begin{figure*}
  \centering
  \includegraphics[width=0.95\textwidth]{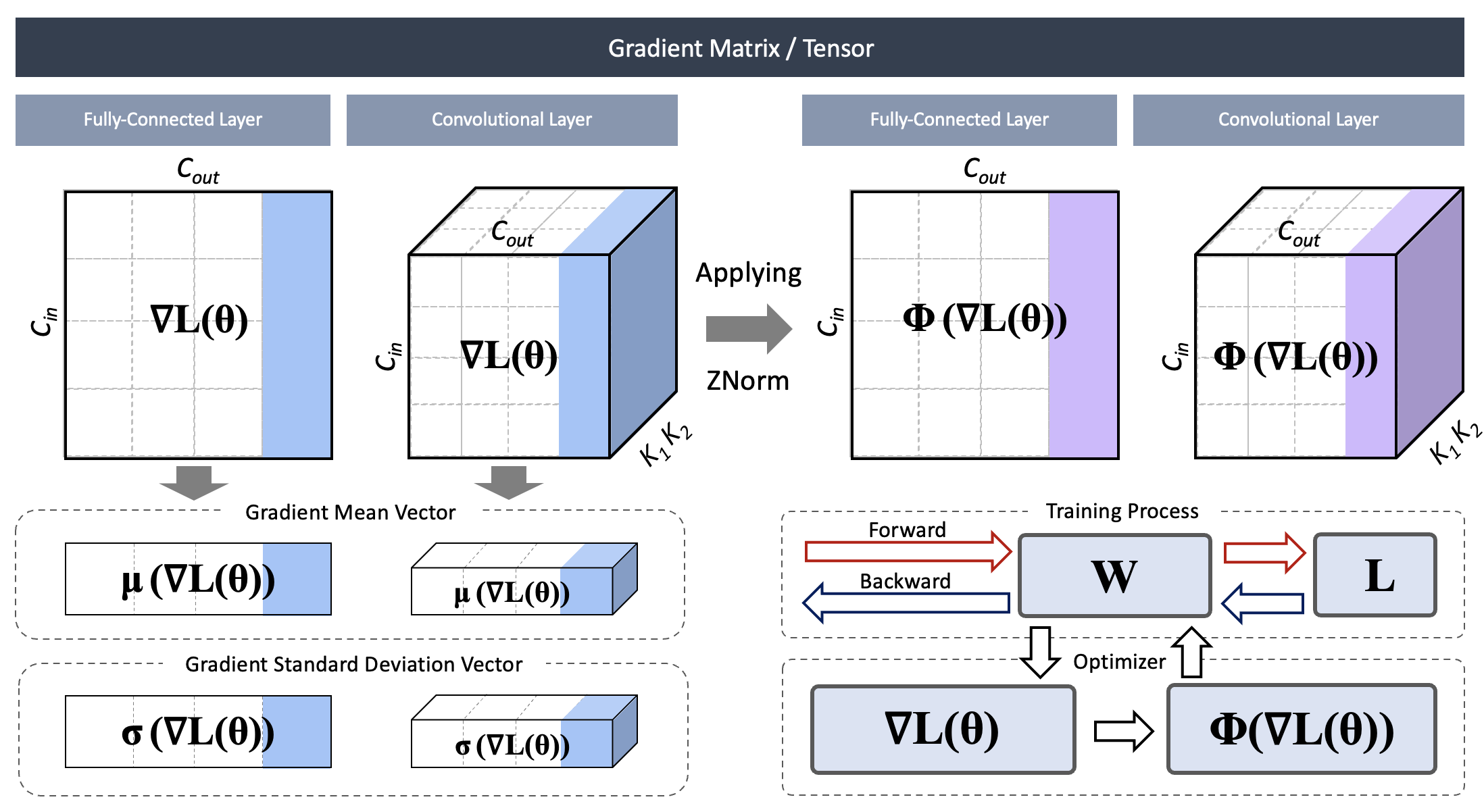}
  \caption{Visualization of the ZNorm($\Phi$) process applied to gradient matrices and tensors in both fully-connected and convolutional layers. The process includes calculating the gradient mean vector and standard deviation vector, followed by normalizing the gradients via ZNorm.  The training process showing how normalized gradients $\Phi(\nabla L(\theta))$ are integrated into the forward-backward optimization loop, with gradient adjustment occurring between gradient computation and the optimizer update step.}
  \label{fig:main}
\end{figure*}

\noindent\textbf{Preliminaries.} Consider a deep neural network consisting of \(L\) layers, where each layer \(l\) has associated weights \(\mathbf{\theta}^{(l)}\). Fully connected layers can be expressed as \(\mathbf{\theta}^{(l)}_{fc} \in \mathbb{R}^{D_l \times M_l}\), where \(D_l\) represents the number of neurons and \(M_l\) represents the dimension of the input to the layer. Convolutional layers, \(\mathbf{\theta}^{(l)}_{conv}\), can be represented as a 4-dimensional tensor \(\mathbf{\theta}^{(l)}_{conv} \in \mathbb{R}^{C_{out}^{(l)} \times C_{in}^{(l)} \times k_1^{(l)} \times k_2^{(l)}}\), where \(C_{in}^{(l)}\) and \(C_{out}^{(l)}\) are the number of input and output channels, respectively, and \(k_1^{(l)}\) and \(k_2^{(l)}\) represent the kernel sizes. Let \(\nabla \mathcal{L}(\mathbf{\theta}^{(l)})\) denote the gradient of the loss function \(\mathcal{L}\) with respect to the weights \(\mathbf{\theta}^{(l)}\) of layer \(l\). \(\nabla \mathcal{L}(\mathbf{\theta})\) denotes the overall gradient tensor. \\

\noindent\textbf{Z-Score Normalization for Gradients.} Z-Score Normalization is applied layer-wise, meaning each layer’s gradient \(\nabla \mathcal{L}(\mathbf{\theta^{(l)}})\) is individually normalized using ZNorm, which can be expressed as:
\begin{equation}
\Phi_{ZNorm}(\nabla \mathcal{L}(\mathbf{\theta_{(t)}^{(l)}})) =  \frac{\nabla \mathcal{L}(\mathbf{\theta_{(t)}^{(l)}}) - \mu_{\nabla \mathcal{L}(\mathbf{\theta_{(t)}^{(l)}})}}{\sigma_{\nabla \mathcal{L}(\mathbf{\theta_{(t)}^{(l)}})} + \epsilon}
\end{equation}
Here, \(\mu_{\nabla \mathcal{L}(\mathbf{\theta^{l}})}\) represents the mean of the gradients in layer \(l\), and \(\sigma_{\nabla \mathcal{L}(\mathbf{\theta^{l}})}\) is their standard deviation. A small constant \(\epsilon\) (typically 1e-10) is added to prevent division by zero. When we refer to \( \Phi_{ZNorm}(\nabla \mathcal{L}(\mathbf{\theta})) \), it denotes the collection of all layer-wise gradients that have been individually normalized by ZNorm across the entire network. ZNorm automatically scales the gradients within each layer without requiring additional hyperparameters.

\begin{algorithm}[H]
\caption{Adam with Z-Score Normalization (ZNorm)}
\textbf{Input:} Weight vector $\theta^0$, step size $\mu$, $\beta_1$, $\beta_2$, $\epsilon$, $\mathbf{m}^0$, $\mathbf{v}^0$ \\
\textbf{Training step:}
\begin{algorithmic}[1]
    \For{$t = 1$ to $T$} 
        \State $\mathbf{g}^t = \nabla \mathcal{L}(\mathbf{\theta})$ \Comment{Compute gradient}
        \State $\hat{\mathbf{g}}^t = \Phi_{ZNorm}(\mathbf{g}^t)$ \Comment{Apply Z-Score Normalization}
        \State $\mathbf{m}^t = \beta_1 \mathbf{m}^{t-1} + (1 - \beta_1)\hat{\mathbf{g}}^t$
        \State $\mathbf{v}^t = \beta_2 \mathbf{v}^{t-1} + (1 - \beta_2)\hat{\mathbf{g}}^t \odot \hat{\mathbf{g}}^t$
        \State $\hat{\mathbf{m}}^t = \frac{\mathbf{m}^t}{1 - \beta_1^t}$
        \State $\hat{\mathbf{v}}^t = \frac{\mathbf{v}^t}{1 - \beta_2^t}$
        \State $\mathbf{\theta}^{t+1} = \mathbf{\theta}^t - \mu \frac{\hat{\mathbf{m}}^t}{\sqrt{\hat{\mathbf{v}}^t} + \epsilon}$
    \EndFor
\end{algorithmic}
\end{algorithm}

\subsection{Embedding ZNorm to Adam}
Z-Score Normalization (ZNorm) can be seamlessly integrated into existing deep neural network (DNN) optimization algorithm, Adam~\cite{kingma2014adam}. ZNorm, applied directly to the gradient tensors. In this paper, we specifically focus on embedding ZNorm into the Adam optimization algorithm, which is one of the most widely used optimization algorithms in deep learning due to its adaptive learning rate properties and effectiveness in training deep models. 

Algorithm 1 shows how ZNorm can be embedded into the Adam algorithm with minimal changes. The ZNorm step is applied immediately after computing the gradient, and then the normalized gradient is utilized in the subsequent Adam updates. \\

\noindent\textbf{Stability of ZNorm Training.} Gradient Centralization~\cite{center} has noted potential instability when applying Z-score standardization directly to gradients. Through our analysis, we identify that this instability primarily stems from the diminishing gradient phenomenon during training convergence. 

First, as training progresses towards convergence, the loss function \(\mathcal{L}(\mathbf{\theta})\) monotonically decreases, leading to progressively smaller gradients. This relationship can be formally expressed as:
\begin{equation}
    \|\nabla \mathcal{L}(\mathbf{\theta}^{(t+1)})\| \leq \|\nabla \mathcal{L}(\mathbf{\theta}^{(t)})\|
\end{equation}
where t denotes the training step~\cite{grad1, grad2, grad4}.

As the model converges to a local minimum \(\mathbf{\theta}^*\), the gradient approaches zero:
\begin{equation}
    \lim_{t \to \infty} \nabla \mathcal{L}(\mathbf{\theta}^{(t)}) = \nabla \mathcal{L}(\mathbf{\theta}^*) = 0
\end{equation}

This convergence behavior creates a critical challenge for Z-score normalization. As gradients become extremely small, their standard deviation \(\sigma_{\nabla \mathbf{\theta}}\) also approaches zero. Consequently, when normalizing the gradients:
\begin{equation}
    \frac{\nabla \mathcal{L}(\mathbf{\theta}) - \mu_{\nabla \mathbf{\theta}}}{\sigma_{\nabla \mathbf{\theta}}}
\end{equation}
the denominator becomes vanishingly small, potentially causing the normalized gradients to explode. Even with a small epsilon term added for numerical stability, this fundamental issue persists near convergence. \\

\noindent\textbf{Role of Skip Connection for ZNorm.} The instability problem mentioned earlier can be resolved by using skip connections~\cite{He2016DeepRL}. Skip connections help prevent gradients from becoming extremely small, thereby maintaining the stability of ZNorm during training. Skip connections are commonly used in various network architectures such as ResNet~\cite{He2016DeepRL}, DenseNet~\cite{Huang2017DenselyCC}, and U-Net~\cite{Unet,Unetp}. Let's consider a deep neural network without skip connections. For a layer \( l \), the gradient of the loss with respect to the weights \(\mathbf{\theta}^{(l)}\) can be expressed as:
\begin{align}
\nabla \mathcal{L}(\mathbf{\theta}^{(l)}) = \frac{\partial \mathcal{L}}{\partial \mathbf{\theta}^{(l)}} = \frac{\partial \mathcal{L}}{\partial \mathbf{z}^{(l)}} \cdot \frac{\partial \mathbf{z}^{(l)}}{\partial \mathbf{\theta}^{(l)}} = \delta^{(l)} \cdot \frac{\partial \mathbf{z}^{(l)}}{\partial \mathbf{\theta}^{(l)}}
\end{align}
where \(\delta^{(l)}\) is the error term at layer \( l \), and \(\mathbf{z}^{(l)}\) is the pre-activation (weighted input) at layer \( l \)~\cite{He2016DeepRL, LeCun2015}. For a deep network, this is recursively expanded using the chain rule:
\begin{align}
    \nabla \mathcal{L}(\mathbf{\theta}^{(l)}) &= \delta^{(L)} \cdot \frac{\partial \mathbf{z}^{(L)}}{\partial \mathbf{z}^{(L-1)}}  \cdots \frac{\partial \mathbf{z}^{(l+1)}}{\partial \mathbf{z}^{(l)}} \cdot \frac{\partial \mathbf{z}^{(l)}}{\partial \mathbf{\theta}^{(l)}} \\
    &= \delta^{(L)} \cdot \prod_{k=l+1}^{L} \frac{\partial \mathbf{z}^{(k)}}{\partial \mathbf{z}^{(k-1)}} \cdot \frac{\partial \mathbf{z}^{(l)}}{\partial \mathbf{\theta}^{(l)}}
\end{align}

Each term \( \frac{\partial \mathbf{z}^{(k+1)}}{\partial \mathbf{z}^{(k)}} \) represents the derivative of the pre-activation of layer \( k+1 \) with respect to the pre-activation of the previous layer \( k \), which can be very small for deep networks, leading to gradient vanishing. 

Consider the inclusion of skip connections in the form:
\begin{equation}
    \mathbf{z}^{(l)} = f(\mathbf{z}^{(l-1)}) + \mathbf{z}^{(l-1)}
\end{equation}
where $f(\mathbf{z}^{(l-1)})$ is the activation function~\cite{He2016DeepRL}. With skip connections, the gradient with respect to $\mathbf{\theta}^{(l)}$ at layer $l$ can be expressed as:
\begin{equation}
\begin{aligned}
    \nabla \mathcal{L}(\mathbf{\theta}^{(l)}) &= \frac{\partial \mathcal{L}}{\partial \mathbf{z}^{(l)}} \cdot \frac{\partial \mathbf{z}^{(l)}}{\partial \mathbf{\theta}^{(l)}} \\
    &= \delta^{(l)} \cdot \frac{\partial}{\partial \mathbf{\theta}^{(l)}} \left( f(\mathbf{z}^{(l-1)}) + \mathbf{z}^{(l-1)} \right) \\
    &= \delta^{(l)} \cdot \left( \frac{\partial f(\mathbf{z}^{(l-1)})}{\partial \mathbf{\theta}^{(l)}} + \frac{\partial \mathbf{z}^{(l-1)}}{\partial \mathbf{\theta}^{(l)}} \right)
\end{aligned}
\end{equation}

Applying the chain rule, we obtain:
\begin{equation}
\begin{aligned}
   \nabla \mathcal{L}(\mathbf{\theta}^{(l)}) &= \frac{\partial \mathcal{L}}{\partial \mathbf{z}^{(L)}} \cdot \prod_{k=l+1}^{L} \left( \frac{\partial f(\mathbf{z}^{(k-1)})}{\partial \mathbf{z}^{(k-1)}} + 1 \right) \cdot \frac{\partial \mathbf{z}^{(l)}}{\partial \mathbf{\theta}^{(l)}} \\
   &= \delta^{(L)} \cdot \prod_{k=l+1}^{L} \left( \frac{\partial \mathbf{z}^{(k)}}{\partial \mathbf{z}^{(k-1)}} + 1 \right) \cdot \frac{\partial \mathbf{z}^{(l)}}{\partial \mathbf{\theta}^{(l)}}
\end{aligned}
\end{equation}

The key observation is that even when $\frac{\partial f(\mathbf{z}^{(l-1)})}{\partial \mathbf{\theta}^{(l)}}$ becomes very small, the additional $1$ term in the product ensures that the gradient $\nabla \mathcal{L}(\mathbf{\theta}^{(l)})$ maintains a non-negligible magnitude~\cite{He2016DeepRL}. This mechanism effectively prevents gradient vanishing and provides stability to the ZNorm training process. The final gradient expression with skip connections can be formally written as:
\begin{equation}
    \nabla \mathcal{L}(\mathbf{\theta}^{(l)}) = \delta^{(L)} \cdot \prod_{k=l+1}^{L} \left( \frac{\partial \mathbf{z}^{(k)}}{\partial \mathbf{z}^{(k-1)}} + 1 \right) \cdot \frac{\partial \mathbf{z}^{(l)}}{\partial \mathbf{\theta}^{(l)}}
\end{equation}
The skip connection effectively keeps the gradient \( \nabla \mathcal{L}(\mathbf{\theta}^{(l)}) \) from becoming zero or extremely small, thereby improving the training stability of ZNorm and ensuring consistent learning progress.


\section{Experimental Results}
In this section, we conducted a series of experiments on both image classification and image segmentation tasks to show the performances of our Z-Score Gradient Normalization (ZNorm) method. \\ 

\begin{table*}
\centering
\renewcommand{\arraystretch}{0.85}
\begin{adjustbox}{width=1\textwidth}
\begin{tabular}{|l|l|lcc|}
\hline
\textbf{Datasets.} & \textbf{Model.} & \textbf{Methods.} & \textbf{Test Accuracy}$\uparrow$ & \textbf{Train Loss} \\
\hline
\hline
\multirow{48}{*}{CIFAR-10~\cite{Krizhevsky2009LearningML}}& \multirow{6}{*}{ResNet-56~\cite{He2016DeepRL}}
& Baseline & 0.802 & 0.0033 \\
& & Gradient Centralization~\cite{center} & 0.804 & 0.0128 \\
&& Gradient Clipping~\cite{clip} & 0.747 & 0.0178 \\
&& Weight Decay 1E-3~\cite{adamw} & 0.798 & 0.0183 \\
&& Weight Decay 1E-4~\cite{adamw} & 0.786 & 0.0125 \\
&& \textbf{ZNorm (Ours)} & \textbf{0.812} & 0.0178 \\
\cline{2-5} 
& \multirow{6}{*}{ResNet-101~\cite{He2016DeepRL}}
& Baseline & 0.770 & 0.0165 \\
&& Gradient Centralization~\cite{center} & 0.813 & 0.0112 \\
&& Gradient Clipping~\cite{clip} & 0.772 & 0.0152 \\
&& Weight Decay 1E-3~\cite{adamw} & 0.812 & 0.0169 \\
&& Weight Decay 1E-4~\cite{adamw} & 0.812 & 0.0167 \\
&& \textbf{ZNorm (Ours)} & \textbf{0.820} & 0.0160 \\
\cline{2-5} 
& \multirow{6}{*}{ResNet-152~\cite{He2016DeepRL}}
& Baseline & 0.795 & 0.0172 \\
&& Gradient Centralization~\cite{center} & 0.797 & 0.0246 \\
&& Gradient Clipping~\cite{clip} & 0.786 & 0.0311 \\
&& Weight Decay 1E-3~\cite{adamw} & 0.776 & 0.0168 \\
&& Weight Decay 1E-4~\cite{adamw} & 0.773 & 0.0212 \\
&& \textbf{ZNorm (Ours)} & \textbf{0.823} & 0.0217 \\
\cline{2-5} 
& \multirow{6}{*}{DenseNet-121~\cite{Huang2017DenselyCC}}
& Baseline & 0.759 & 0.0120 \\
&& Gradient Centralization~\cite{center} & 0.784 & 0.0186 \\
&& Gradient Clipping~\cite{clip} & 0.765 & 0.0142 \\
&& Weight Decay 1E-3~\cite{adamw} & 0.776 & 0.0222 \\
&& Weight Decay 1E-4~\cite{adamw} & 0.774 & 0.0108 \\
&& \textbf{ZNorm (Ours)} & \textbf{0.799} & 0.0139 \\
\cline{2-5} 
& \multirow{6}{*}{DenseNet-169~\cite{Huang2017DenselyCC}}
& Baseline & 0.766 & 0.0243 \\
&& Gradient Centralization~\cite{center} & 0.787 & 0.0121 \\
&& Gradient Clipping~\cite{clip} & 0.767 & 0.0144 \\
&& Weight Decay 1E-3~\cite{adamw} & 0.780 & 0.0070 \\
&& Weight Decay 1E-4~\cite{adamw} & 0.798 & 0.0002 \\
&& \textbf{ZNorm (Ours)} & \textbf{0.802} & 0.0150 \\
\cline{2-5} 
& \multirow{6}{*}{MobileNetV2~\cite{sandler2018mobilenetv2}}
& Baseline & 0.763 & 0.0242 \\
&& Gradient Centralization~\cite{center} & 0.761 & 0.0282 \\
&& Gradient Clipping~\cite{clip} & 0.775 & 0.0266 \\
&& Weight Decay 1E-3~\cite{adamw} & 0.742 & 0.0232 \\
&& Weight Decay 1E-4~\cite{adamw} & 0.757 & 0.0276 \\
&& \textbf{ZNorm (Ours)} & \textbf{0.780} & 0.0332 \\
\cline{2-5} 
& \multirow{6}{*}{VGG-16~\cite{Simonyan2015VeryDC}}
& \textbf{Baseline} & \textbf{0.845} & 0.0138 \\
&& Gradient Centralization~\cite{center} & 0.837 & 0.0167 \\
&& Gradient Clipping~\cite{clip} & 0.844 & 0.0149 \\
&& Weight Decay 1E-3~\cite{adamw} & 0.821 & 0.0148 \\
&& Weight Decay 1E-4~\cite{adamw} & 0.841 & 0.0153 \\
&& ZNorm (Ours) & 0.835 & 0.0253 \\
\cline{2-5} 
& \multirow{6}{*}{Xception~\cite{Chollet2017XceptionDL}}
& \textbf{Baseline} & \textbf{0.751} & 0.0160 \\
&& Gradient Centralization~\cite{center} & 0.740 & 0.0160 \\
&& Gradient Clipping~\cite{clip} &  0.741 & 0.0194 \\
&& Weight Decay 1E-3~\cite{adamw} & 0.739 & 0.0221 \\
&& Weight Decay 1E-4~\cite{adamw} & 0.737 & 0.0184 \\
&& ZNorm (Ours) & 0.728 & 0.0562 \\
\hline
\end{tabular}
\label{tab:network}
\end{adjustbox}
\caption{
Performance Comparison on CIFAR-10 Dataset~\cite{Krizhevsky2009LearningML} for various convolutional neural network architectures and gradient normalization techniques. The table shows the test accuracy and train loss across several models and normalization methods. Bold values represent the highest test accuracy achieved for each model. ZNorm consistently outperforms other methods in terms of test accuracy, demonstrating its effectiveness in improving model performance.
}
\end{table*}

\begin{table*}[htbp]
\centering
\renewcommand{\arraystretch}{0.85}
\begin{adjustbox}{width=1\textwidth}
\begin{tabular}{|l|l|lcc|}
\hline
\textbf{Datasets.} & \textbf{Model.} & \textbf{Methods.} & \textbf{Test Accuracy}$\uparrow$ & \textbf{Train Loss} \\
\hline
\hline

& \multirow{6}{*}{ResNet-56~\cite{He2016DeepRL}}
& Baseline & 0.880 & 0.2349 \\
& & Gradient Centralization~\cite{center} & 0.887 & 0.0252 \\
& & Gradient Clipping~\cite{clip} & 0.887 & 0.0300 \\
&& Weight Decay 1E-3~\cite{adamw} & 0.847 & 0.3085 \\
&& Weight Decay 1E-4~\cite{adamw} & 0.875 & 0.0121 \\
& &\textbf{ ZNorm (Ours)} & \textbf{0.915} & 0.0418 \\
\cline{2-5} 
& \multirow{6}{*}{ResNet-101~\cite{He2016DeepRL}}
& Baseline & 0.870 & 0.1012 \\
& & Gradient Centralization~\cite{center} & 0.910 & 0.1241 \\
PatchCamelyon~\cite{pcam}  & & Gradient Clipping~\cite{clip} & 0.915 & 0.1004 \\
(2) && Weight Decay 1E-3~\cite{adamw} & 0.845 & 0.3213 \\
&& Weight Decay 1E-4~\cite{adamw} & 0.839 & 0.1511 \\
& &\textbf{ ZNorm (Ours) }& \textbf{0.917} & 0.1322 \\
\cline{2-5} 
& \multirow{6}{*}{ResNet-152~\cite{He2016DeepRL}}
& Baseline & 0.884 & 0.0504 \\
& & Gradient Centralization~\cite{center} & 0.882 & 0.0695 \\
& & Gradient Clipping~\cite{clip} & 0.910 & 0.1231 \\
&& Weight Decay 1E-3~\cite{adamw} & 0.902 & 0.0537 \\
&& Weight Decay 1E-4~\cite{adamw} & 0.882 & 0.1021 \\
& & \textbf{ZNorm (Ours)} & \textbf{0.912} & 0.0591 \\
\hline
\end{tabular}
\label{tab:tumor}
\end{adjustbox}
\caption{Performance comparison of Breast Tumor Prediction on PatchCamelyon~\cite{pcam} dataset based on ResNet~\cite{He2016DeepRL} models with different normalization techniques. Bold values represent the highest test accuracy achieved for each model. ZNorm consistently outperforms other methods in terms of test accuracy, demonstrating its effectiveness in improving model performance}
\end{table*}

\begin{table*}[htbp]
\centering
\renewcommand{\arraystretch}{0.85}
\begin{adjustbox}{width=1\textwidth}
\begin{tabular}{|l|l|lccc|}
\hline
\textbf{Datasets.} & \textbf{Model.} & \textbf{Methods.} & \textbf{Test F1}$\uparrow$ & \textbf{Test Tversky}$\uparrow$ & \textbf{Test Hausdorff Dist}$\downarrow$ \\
\hline
\hline

& \multirow{6}{*}{ResNet50-Unet~\cite{Unet}}
& Baseline & 0.901 & 0.881 & 2.767 \\
&& Gradient Centralization~\cite{center} & 0.893 & 0.872 & 2.834 \\
&& Gradient Clipping~\cite{clip} & 0.883 & 0.882 & 2.971 \\
&& Weight Decay 1E-3~\cite{adamw} & 0.904 & 0.866 & 2.762\\
&& Weight Decay 1E-4~\cite{adamw} & 0.904 & 0.894 & 2.762\\
&& \textbf{ZNorm (Ours)} & \textbf{0.917} & \textbf{0.914} & \textbf{2.663} \\
\cline{2-6} 
& \multirow{6}{*}{U-net++~\cite{Unetp}}
& Baseline & 0.881 & 0.901 & 3.001 \\
 & & Gradient Centralization~\cite{center} & \textbf{0.901} & 0.899 & \textbf{2.822} \\
LGG MRI & & Gradient Clipping~\cite{clip} & 0.896 & 0.902 & 2.871 \\
Dataset~\cite{tumordata2} && Weight Decay 1E-3~\cite{adamw} & 0.897 & 0.903 & 2.890\\
&& Weight Decay 1E-4~\cite{adamw} & 0.870 & 0.889 & 3.063\\
&& \textbf{ZNorm (Ours)} & 0.898 & \textbf{0.910} & 2.840 \\
\cline{2-6} 
& \multirow{6}{*}{Attention U-Net~\cite{AttentionUnet}}
& Baseline & 0.867 & 0.912 & 3.162 \\
&& Gradient Centralization~\cite{center} & 0.864 & 0.916 & 3.135 \\
&& Gradient Clipping~\cite{clip} & 0.878 & 0.926 & 3.039\\
&& Weight Decay 1E-3~\cite{adamw} & 0.878 & 0.909 & 3.053\\
&& Weight Decay 1E-4~\cite{adamw} & 0.871 & 0.908 & 3.159\\
&& \textbf{ZNorm (Ours)} & \textbf{0.894} & \textbf{0.928} & \textbf{2.947} \\
\hline
\end{tabular}
\label{tab:tumorseg}
\end{adjustbox}
\caption{Performance Comparison of Brain Tumor Segmentation on LGG MRI Dataset~\cite{tumordata2} with various architectures and normalization methods. Bold values represent the highest test accuracy achieved for each model. ZNorm consistently outperforms other methods in terms of Test Tversky, demonstrating its effectiveness in improving model performance}
\end{table*}

\noindent\textbf{Experimental Settings.} All experiments were performed using the Adam optimizer~\cite{kingma2014adam}, with the baseline referring to the standard Adam gradient. The experiments involved Gradient Clipping~\cite{clip}, Gradient Centralization~\cite{center}, and ZNorm (Ours). Additionally, Weight Decay 3 and Weight Decay 4 correspond to decay rates of 1E-3 and 1E-4, respectively. For clipping, the value was 0.1. For image classification, we used the CIFAR-10~\cite{Krizhevsky2009LearningML} and PatchCamelyon~\cite{pcam} datasets. All datasets were not augmented. All architectures used in our experiments (ResNet, DenseNet, VGG-16, U-Net variants) include batch normalization layers by default.

Both were trained with a batch size of 256, a learning rate of 0.001, and 100 epochs for CIFAR-10 and 50 epochs for PatchCamelyon. For image segmentation, we used the LGG MRI dataset~\cite{tumordata2} for brain tumor segmentation. Training settings were a batch size of 128, an initial learning rate of 0.01, and total training for 50 epochs, with the learning rate reduced by a factor of 10 every 5 epochs starting from epoch 30. \\

\noindent\textbf{Evaluation Metric: Hausdorff Distance.} The Hausdorff Distance is a metric used to evaluate the maximum discrepancy between two sets of points, commonly employed in segmentation tasks to measure the spatial alignment between the predicted segmentation mask and the ground truth mask. Specifically, it quantifies the greatest distance from a point in one set to the closest point in the other set. For two sets of points $P$ (predicted mask) and $G$ (ground truth mask), the Hausdorff Distance is defined as:
\begin{align}
    H(P, G) = \max\{\sup_{p \in P} \inf_{g \in G} d(p, g), \sup_{g \in G} \inf_{p \in P} d(g, p)\}
\end{align}
where $d(p, g)$ is the Euclidean distance between point $p \in P$ and point $g \in G$. The metric calculates the maximum of the minimum distances between the two sets, capturing the largest deviation in their alignment. In segmentation tasks, a lower Hausdorff Distance indicates better spatial alignment between the predicted mask and the ground truth mask, highlighting the precision of the segmentation model in delineating boundaries.

\subsection{Image Classification Results for CIFAR-10}

In this section, we assess the effectiveness of ZNorm on the CIFAR-10 dataset~\cite{Krizhevsky2009LearningML} across various neural network architectures. ZNorm consistently improves performance across different CNN architectures, particularly in skip-connected networks. Table 1 provides detailed experimental results, showing how ZNorm enhances performance across multiple architectures. For instance, in ResNet-56~\cite{He2016DeepRL}, ZNorm achieves a test accuracy of 0.812, outperforming the baseline (0.802) as well as methods like Gradient Centralization (0.804) and Gradient Clipping (0.747). Similarly, ResNet-101~\cite{He2016DeepRL} shows improved results with ZNorm, reaching a test accuracy of 0.820, surpassing the baseline (0.770) and other techniques. In DenseNet-121~\cite{Huang2017DenselyCC}, ZNorm boosts test accuracy to 0.799, outperforming the baseline (0.759) and Gradient Centralization (0.784). DenseNet-169~\cite{Huang2017DenselyCC} also benefits from ZNorm, achieving the highest test accuracy of 0.802, compared to the baseline (0.766).
However, in networks without skip connections, such as VGG-16~\cite{Simonyan2015VeryDC} and Xception~\cite{Chollet2017XceptionDL}, ZNorm shows comparable but not superior performance (0.835 vs 0.845 for VGG-16, 0.728 vs 0.751 for Xception). These results demonstrate that ZNorm is particularly effective in improving accuracy for skip-connected architectures, where maintaining stable gradient flow is crucial for performance.

\subsection{Image Classification Results for Breast Tumor}
We tested ZNorm on the PatchCamelyon dataset~\cite{pcam}, which contains breast cancer images, to check if it works well on real medical data. Using three different ResNet models (ResNet-56, -101, and -152)~\cite{He2016DeepRL}, our results in Table 2 show that ZNorm works better than other methods. ZNorm achieved high test accuracy across all models: 0.915 with ResNet-56 (compared to baseline 0.880), 0.917 with ResNet-101 (baseline 0.870), and 0.912 with ResNet-152 (baseline 0.884). These results show that ZNorm works well for real medical tasks like finding tumors in breast cancer images.
\subsection{Image Segmentation for Brain Tumor}
\begin{figure*}
  \centering
  \includegraphics[width=1\textwidth]{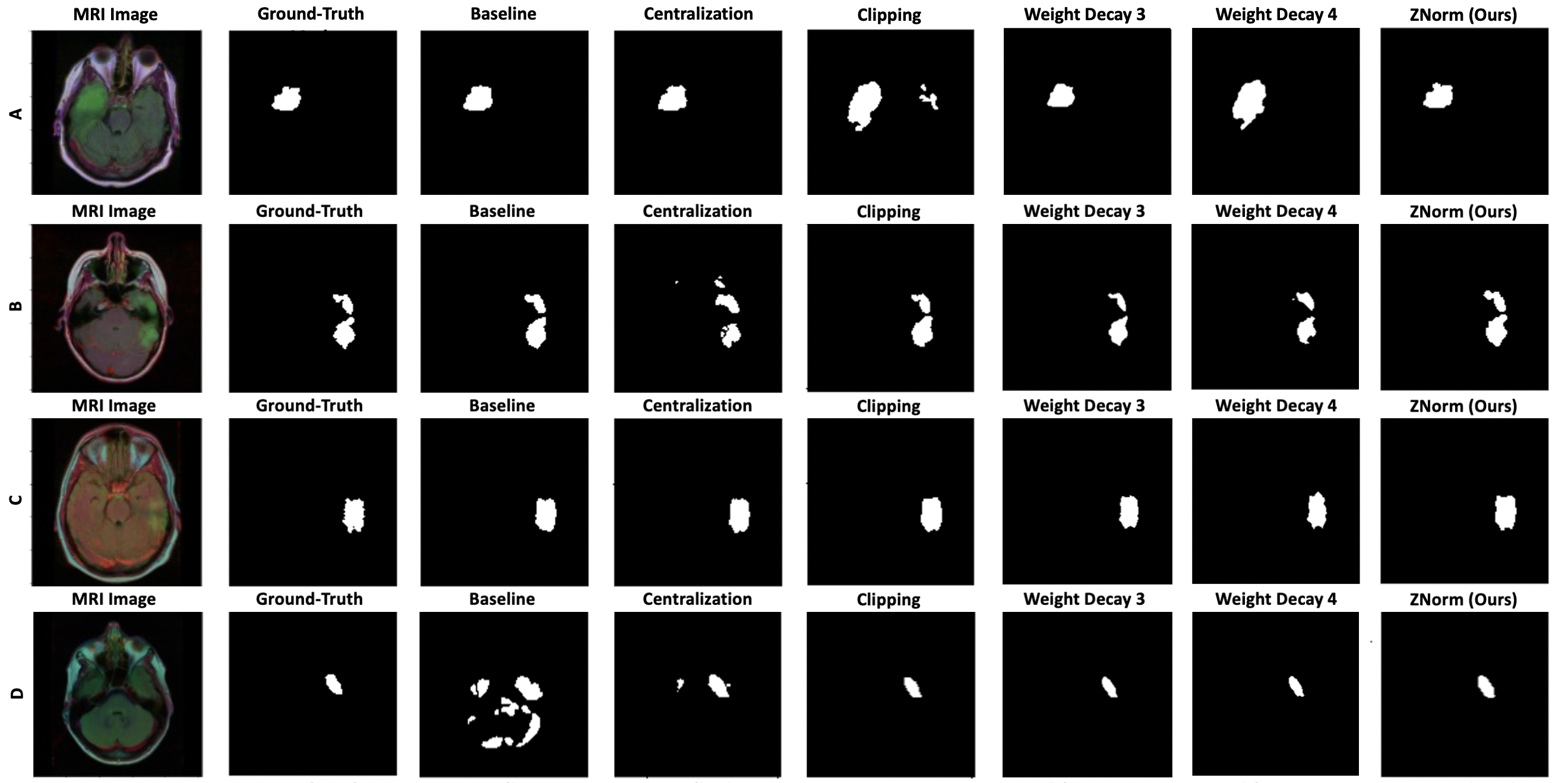}
  \caption{
  Comparison of segmentation mask results using different methods such as GC\cite{center}, Clipping\cite{clip}, Weight Decays~\cite{adamw} and ZNorm on LGG datasets\cite{tumordata2} based on ResNet-50-Unet~\cite{Unet}. ZNorm demonstrates superior performance, producing segmentation masks that more closely match the ground-truth compared to other methods.
  }
  \label{fig:segment}
\end{figure*}

For the brain tumor segmentation task, we tested the LGG MRI dataset~\cite{tumordata2} using three skip-connected models: ResNet50-Unet~\cite{Unet}, U-net++\cite{Unetp}, and Attention U-Net\cite{AttentionUnet}. These models use skip connections, which help ZNorm work better as explained earlier. Looking at Table 3, ZNorm does better than other methods across all models. Using ResNet50-Unet, ZNorm got the best test F1 score (0.917) and Tversky score (0.914), and the smallest Hausdorff distance (2.663). For U-net++, while Gradient Centralization had a slightly better F1 score (0.901), ZNorm still did well (F1: 0.898, Tversky: 0.910). With Attention U-Net, ZNorm got the best Tversky score (0.928) and smallest Hausdorff distance (2.947). These results show that ZNorm makes tumor segmentation better across different models, especially by reducing Hausdorff distance while keeping good F1 and Tversky scores. Looking at the mask in Figure~\ref{fig:segment}, we can see that ZNorm makes tumor outlines that match the real tumors better than other methods like Gradient Centralization~\cite{center}, Gradient Clipping~\cite{clip}, and different weight decay settings~\cite{adamw}.

\section{Discussion}

\subsection{Impact on Skip-Connected Architectures.}
The role of skip connections in ensuring the stability of ZNorm was particularly evident in our results. By preserving gradient flow and preventing vanishing gradients, skip connections in architectures like ResNet and U-Net enhanced ZNorm's ability to stabilize training. This synergy between ZNorm and skip connections highlights its suitability for these architectures, making it a valuable tool for efficiently training networks that leverage residual and dense connections.

\subsection{Practical Implications}
ZNorm's simplicity and ease of implementation make it a practical choice for real-world applications. Unlike normalization methods that require architectural modifications or additional hyperparameters, ZNorm operates directly on gradients, allowing it to be seamlessly integrated into existing optimization pipelines. This adaptability reduces the computational and developmental overhead typically associated with implementing new training methods, making ZNorm particularly appealing for applications requiring rapid prototyping and deployment.

\subsection{Limitations and Future Research}
While our study demonstrates the effectiveness of ZNorm across various tasks and architectures, it also highlights several directions for extended research. One limitation is that the experiments were primarily conducted on relatively small- to medium-scale datasets such as CIFAR-10 and medical datasets like PatchCamelyon and LGG MRI. These datasets effectively showcase the benefits of ZNorm, but future research will focus on larger and more diverse datasets, such as ImageNet, to validate its scalability and generalizability. Conducting experiments on ImageNet will provide critical insights into whether the performance improvements observed in this study remain consistent in larger-scale and more complex tasks.

Another area for exploration involves addressing the challenges ZNorm may encounter near convergence. As gradient magnitudes become extremely small during this phase, there is a risk of instability in the normalization process. Although the use of skip connections mitigates this issue to some extent, future work will explore adaptive mechanisms to further stabilize ZNorm, ensuring robustness across all stages of training.

These extensions will build upon the findings of this study, offering opportunities to further enhance the training efficiency, stability, and scalability of ZNorm across an even broader range of applications and domains. By addressing these areas, we aim to establish ZNorm as a foundational tool for efficient deep learning in increasingly complex and diverse settings

\section{Conclusion}
In this paper, we proposed Z-score Gradient Normalization (ZNorm), a novel and efficient method for training skip-connected neural networks. By normalizing both the mean and variance of gradients, ZNorm mitigates common challenges such as vanishing and exploding gradients without requiring architectural modifications. This simplicity ensures seamless integration into existing training pipelines while enhancing the training stability and performance of deep neural networks. Our experimental results demonstrate the effectiveness of ZNorm across diverse tasks and datasets. On image classification tasks with CIFAR-10 and PatchCamelyon, ZNorm consistently outperformed baseline methods and other gradient adjustment techniques. In medical image segmentation, ZNorm achieved superior results on the LGG MRI dataset, particularly in reducing the Hausdorff Distance while maintaining high Tversky and F1 scores. These results highlight ZNorm's capability to improve performance and generalization in both standard and high-impact applications, especially in skip-connected architectures like ResNet and U-Net. In conclusion, ZNorm offers a robust and straightforward approach to optimizing deep neural network training. Its ability to enhance performance without modifying network architecture positions it as a valuable tool for advancing efficient training methods in deep learning. Future work will build upon these findings to refine and expand ZNorm's applicability in the evolving landscape of deep learning research.

\bibliographystyle{plain}
\bibliography{main}

\end{document}